\pdfoutput=1

\documentclass[11pt]{article}

\usepackage[]{ACL2023}

\usepackage{times}
\usepackage{latexsym}

\usepackage[T1]{fontenc}

\usepackage[utf8]{inputenc}

\usepackage{microtype}

\usepackage{inconsolata}

\usepackage{svg}

\usepackage{graphicx} 

\usepackage{multirow}

\usepackage{tabularx}

\usepackage{booktabs}
\newcommand{\ra}[1]{\renewcommand{\arraystretch}{#1}}

\usepackage{varwidth}

\newcommand{\gpt}{\texttt{GPT3-D3} }
\newcommand{\gptwithoutspace}{\texttt{GPT3-D3}}

\usepackage[textsize=scriptsize]{todonotes}

\title{Summarizing, Simplifying, and Synthesizing Medical Evidence Using GPT-3 (with Varying Success)}

\author{
\quad\textbf{Chantal Shaib$^1$}\quad\quad\ \ 
\textbf{Millicent L. Li$^1$}\quad\quad
\textbf{Sebastian Joseph$^2$}\quad\quad\\
\textbf{Iain J. Marshall$^3$}\quad\quad
\textbf{Junyi Jessy Li$^2$}\quad\quad
\textbf{Byron C. Wallace$^1$}\\
$^1$Northeastern University, $^2$The University of Texas at Austin, \\ $^3$King's College London \\
\small\texttt{\{shaib.c, li.mil, b.wallace\}@northeastern.edu} \\
\small\texttt{iain.marshall@kcl.ac.uk} \\
\small\texttt{\{sebaj, jessy\}@utexas.edu} \\
}

\begin{document}
\maketitle
\begin{abstract}
Large language models, particularly GPT-3, are able to produce high quality summaries of general domain news articles in few- and zero-shot settings 
However, it is unclear if such models are similarly capable in more specialized, high-stakes domains such as biomedicine. In this paper, we enlist domain experts (individuals with medical training) to evaluate summaries of biomedical articles generated by GPT-3, given zero supervision. We consider both single- and multi-document settings. In the former, GPT-3 is tasked with generating regular and plain-language summaries of articles describing randomized controlled trials; in the latter, we assess the degree to which GPT-3 is able to \emph{synthesize} evidence reported across a collection of articles. We design an annotation scheme for evaluating model outputs, with an emphasis on assessing the factual accuracy of generated summaries. We find that while GPT-3 is able to summarize and simplify single biomedical articles faithfully, it struggles to provide accurate aggregations of findings over multiple documents. We release all data and annotations used in this work.\footnote{\url{https://github.com/cshaib/summarizing-medical-evidence}}
\end{abstract}

\section{Introduction}

\begin{figure}
    \centering
    \includegraphics[scale=0.375]{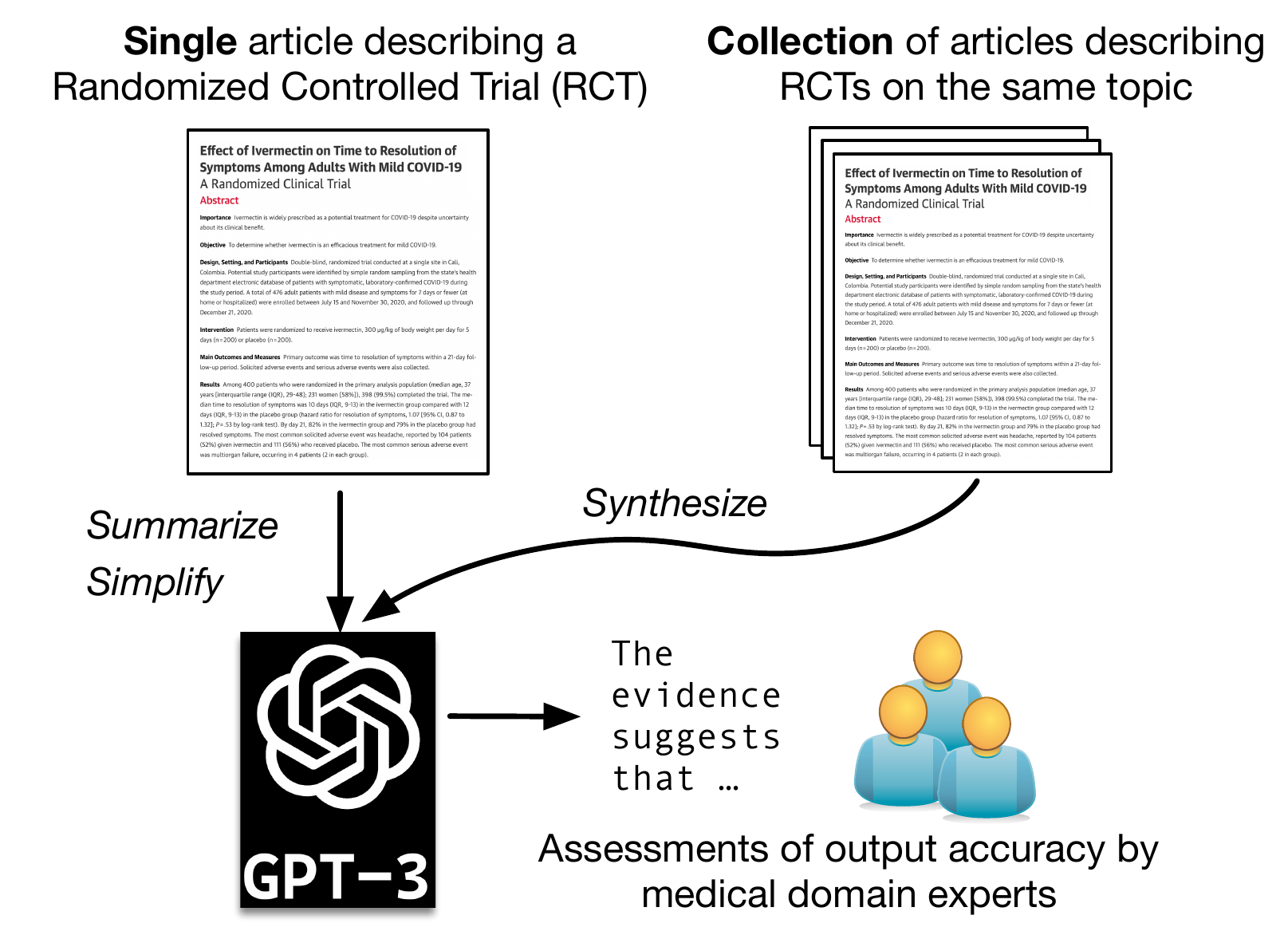}
    \caption{We enlist domain experts to evaluate the factual accuracy of summaries and simplifications of medical articles describing clinical trials. We consider both single- and multi-document settings.}
    \label{fig:overview}
\end{figure}

Large language models have been shown to be capable of producing high-quality and reasonably accurate summaries in \emph{zero-shot} settings \cite{Goyal2022NewsSA, liang2022holistic}, 
with GPT-3 besting fully supervised models in generic news summarization, according to human judgments \cite{Goyal2022NewsSA}. 
In this work we evaluate if such models are similarly 
able to summarize medical literature, a high-stakes domain that demands factual accuracy. 

Specifically, we use the newest iteration of GPT-3 (\texttt{text-davinci-003}; \gpt from here) to generate summaries of (a) individual articles describing individual randomized controlled trials (RCTs) evaluating the efficacy of interventions, and, (b) collections of such articles that describe several trials addressing the same underlying clinical question (e.g., evaluating the same medication). 
These constitute single- and multi-document summarization tasks, respectively. In the single-document case, we also evaluate the ability of \gpt to summarize in \emph{plain language}. We enlist domain experts (with medical training) to annotate model outputs, and seek to address the following questions. 


\vspace{0.2em}
\noindent {\bf RQ1} Does \gpt produce \emph{faithful} summaries of medical articles? 

\vspace{0.2em}
\noindent {\bf RQ2} Can \gpt accurately \emph{simplify} while also summarizing such texts?

\vspace{0.2em}
\noindent {\bf RQ3} Can \gpt \emph{synthesize}---aggregate the findings presented in---multiple input articles in a way that accurately reflects the totality of the evidence?

\vspace{0.2em}
\noindent {\bf RQ4} What sort of factual mistakes does \gpt make when performing these tasks (if any), and what are the risks implied by such errors? 

Overall, we find that \gpt performs single-document summarization and simplification with reasonably good accuracy. 
However, it is less able to accurately synthesize evidence reported in \emph{collections} of trials (in the multi-document case). We release all model outputs and accompanying annotations to facilitate additional work on this topic. 


\section{Single Document Summarization}
\label{sec:prompts}

\paragraph{Data}
We sample 100 articles describing randomized control trials (RCTs) indexed in the Trialstreamer database \cite{marshall2020trialstreamer}, which also provides automatically extracted ``key results''\footnote{ Extracted sentence communicating the main findings.} alongside titles and abstracts. We search for trials published after November 28 2022, following the release date of \gptwithoutspace, to ensure the model has not seen any of the studies during pre-training. 

\vspace{-0.3em}
\paragraph{Experimental Setup}
Using the RCT data described above, we evaluate the ability of \gpt to faithfully summarize and simplify biomedical texts in a zero-shot setting. We also compare \gpt summaries to summaries generated using Flan-T5 \cite{Wei2021FinetunedLM}, but qualitatively find that \gpt summaries are much higher quality. We provide results of this comparison in Appendix \ref{flan}.
Specifically, we prompt \gpt to separately produce: (i) a technical summary, and, (ii) a plain language summary \cite{August2022PaperPM}. See Appendix \ref{sec:appendix_prompts} for all prompts.

\vspace{-0.3em}
\paragraph{Study Design}
\label{sec:studydesign}

We designed an evaluation scheme that captures the sensitivity of medical information.
To assess factuality, we collect annotations about omissions and errors with respect to main results, and key components of the trials including populations, interventions, and outcomes (``PICO'' elements; \citealt{richardson1995well}). 
Where appropriate, we ask annotators to highlight spans of generated text that are inconsistent with the input---these might be ``new'' concepts introduced or spans that directly contradict the input. 
To gauge overall linguistic quality, we solicit assessments regarding the fluency and usefulness of a summary on a Likert scale (\citeyear{likert1932technique}). 
We include additional questions about the simplification of technical terms for the plain language summaries. 
We provide a complete taxonomy of the survey in Appendix \ref{sec:survey}.

\vspace{-0.3em}
\paragraph{Annotations} We recruited 3 domain experts with medical training on the Upwork platform,\footnote{\url{https://www.upwork.com}} and task them each with annotating 100 samples. In total, we collect 300 annotations (3 annotations per sample). 
We use Label Studio\footnote{\url{https://labelstud.io/}} as our interface.

\section{Multiple Document Summarization and Evidence Synthesis} 

\paragraph{Data}
For multi-document summarization, we download  meta-analyses from the Cochrane Library (these are reviews of medical evidence, usually RCTs).\footnote{\url{https://www.cochranelibrary.com/}} 
Our final sample contains 50 multi-document studies comprising meta-review titles, reference abstracts (inputs), and target conclusions (target summaries) 
written by domain experts, 10 of which were published post-\gpt release. \footnote{At the time of retrieval we were only able to extract 18 samples post-\gpt release. We excluded any updates (meta-analyses with $\leq$ 1 reference abstract). There was no discernible difference in the performance, however,  more data is needed to evaluate this effect}

\vspace{-0.3em}
\paragraph{Experimental Setup}
Because inputs comprise multiple abstracts, these (together with generated tokens) often exceed the token capacity of \gptwithoutspace. In our dataset, about 41\% of the samples exceeded this upper-bound. We report information about our data, including average length, in Appendix \ref{sec:dataset_statistics}. To address the upper-bound problem, we adopt a simple two-phase strategy for multi-document summarization. First, we  generate independent summaries for each abstract, using the single-document summarization prompt described in Section \ref{sec:prompts}. Then, we include all the generated single-document summaries in our multi-document synthesis prompt\footnote{Note that 
we have yet to see prior work systematically investigate a strategy for zero-shot multi-document summarization; 
due to the prompt-sensitive nature of LLMs~\cite{liang2022holistic}, we do not guarantee that we obtained the best prompt despite fairly extensive trials.
} (examples in Appendix \ref{sec:appendix_prompts}).

\vspace{-0.3em}
\paragraph{Study Design}
Our evaluation rubric asks for assessments of generated outputs as compared to: (a) inputs, and, (b) target summaries. Specifically, we ask if generated summaries are supported by the \emph{summaries} provided as inputs in the multi-document case, and to what extent they agree with target (reference) summaries. We also ask annotators to highlight spans of text in generated outputs that disagree with paired target summaries. 
We reproduce the full rubric in Appendix \ref{sec:survey}.

With respect to annotators, we use the same procedure described in Section \ref{sec:studydesign}; we recruited 3 new medical experts and tasked them each with annotating 50 samples, for a total of 150 annotations. 

\section{Results}

\begin{figure}[tb!] 
    \centering
    \includegraphics[width=0.45\textwidth]{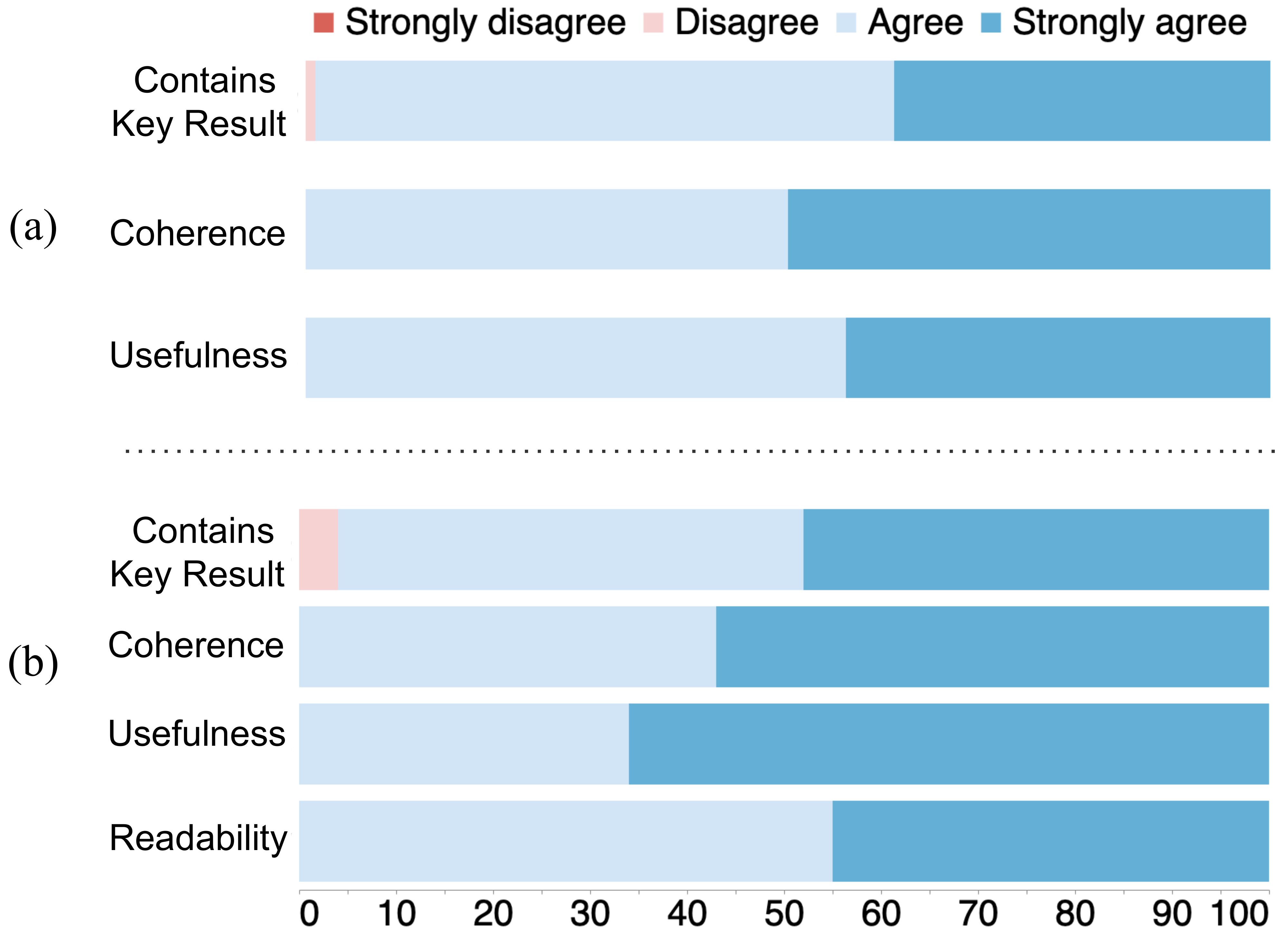}
    \vspace{-0.7em}
    \caption{Average scores for assessing overall faithfulness, coherence, and usefulness of generated (a) regular summaries and (b) simplified summaries. \gpt produces high-quality regular and simplified summaries.}
    \label{fig:fig2}
\end{figure}

\paragraph{RQ1: Does \gpt produce faithful summaries of medical articles?}
In the single document setting, we find that \gpt generates summaries of biomedical abstracts that are fairly high-quality. Figure \ref{fig:fig2} (a) shows that annotators rated a majority of the summaries as being coherent, useful, and capturing ``key results''.

\begin{figure}
    \centering
    \includegraphics[width=0.45\textwidth]{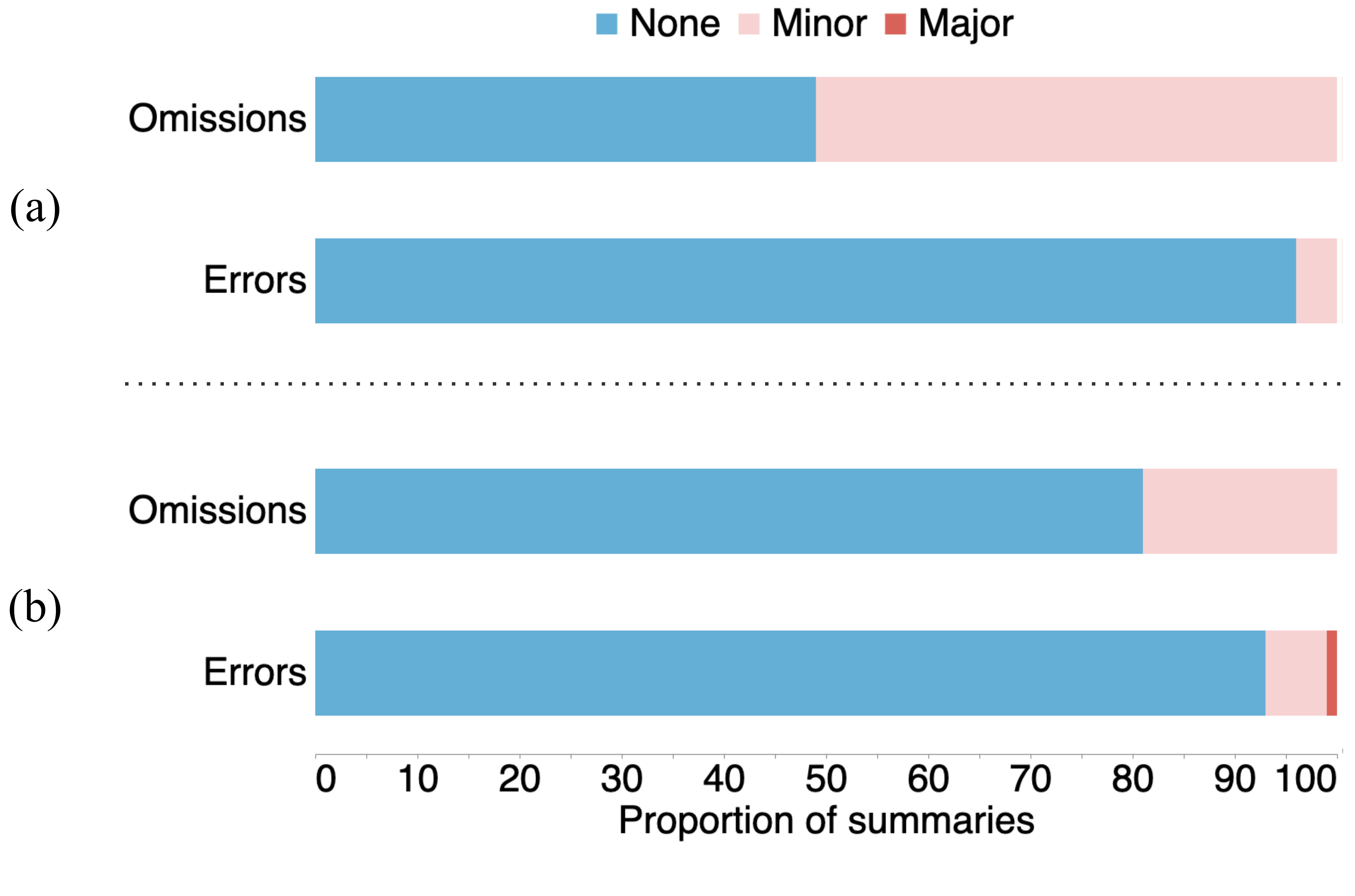}
    \vspace{-0.7em}
    \caption{Average number of errors and omissions made in the generated (a) regular and (b) simplified summaries. Most mistakes made in both cases are minor, and omissions are more frequent than errors.}
    \label{fig:fig3}
\end{figure}

When \gpt does err, it tends to make minor mistakes or omit details. The latter is more common than the former, as shown in Figure \ref{fig:fig3} (a). 

\paragraph{RQ2: Can \gpt accurately simplify while summarizing medical texts?}
Shown in Figure \ref{fig:fig2} (b), \gpt produces simplified summaries that are similarly deemed to be coherent and useful, and which appear to contain key results.
Simplified outputs are scored highly in terms of readability, indicating that these summaries would be understood by someone without medical training.  

In comparison to the technical summaries, Figure \ref{fig:fig3} (b) shows that there are fewer omissions but a slightly higher amount of errors. 
These may be problematic, but --- importantly --- some omissions are expected in a simplified summary, as certain details that are important for an accurate summary for a technical audience may not be necessary to convey key information to a more general audience.   

\begin{figure}
    \centering
\includegraphics[width=0.45\textwidth]{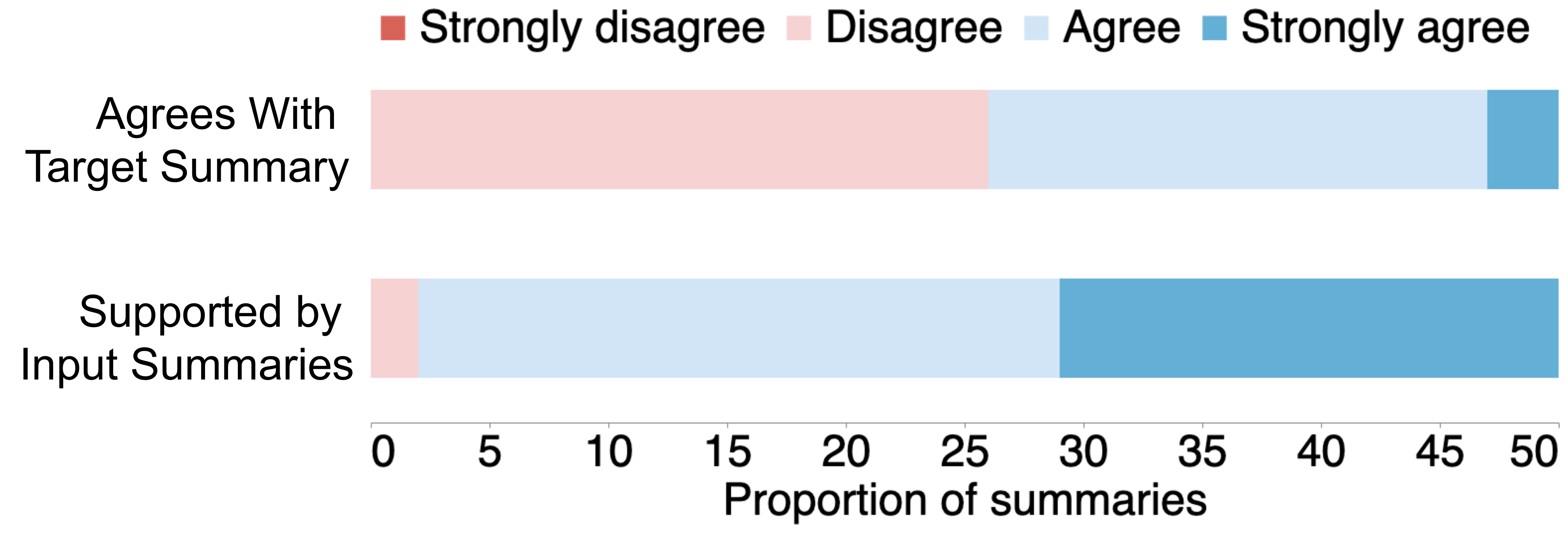}
    \vspace{-0.3em}
    \caption{Proportion of summaries that reflect the target summary and are supported by the input summaries in the multi-document setting. While most summaries follow from the input, less than half are rated as agreeing with the target summary.}
    \label{fig:fig4}
\end{figure}

\paragraph{RQ3: Can \gpt \emph{synthesize} findings presented in multiple input articles in a way that accurately reflects the totality of the evidence?}
We now evaluate \gptwithoutspace's performance on multi-document summarization, i.e., its ability to synthesize evidence \cite{wang-etal-2022-overview}.  
Figure \ref{fig:fig4} shows that most summaries generated by \gpt in this setting are supported by the inputs. 
This is consistent with our findings in \textbf{RQ1}: \gpt is able to summarize faithfully with respect to given input. 
However, we find that generated summaries do not consistently agree with the target summaries. 
Indeed, Figure \ref{fig:fig4} shows that generated summaries disagree with the targets in over half of cases.
This discrepancy suggests that human-written summaries in the biomedical domain require a level of synthesis that is not captured by \gpt. 


\begin{figure}[tb!]
    \centering
    \includegraphics[width=0.5\textwidth]{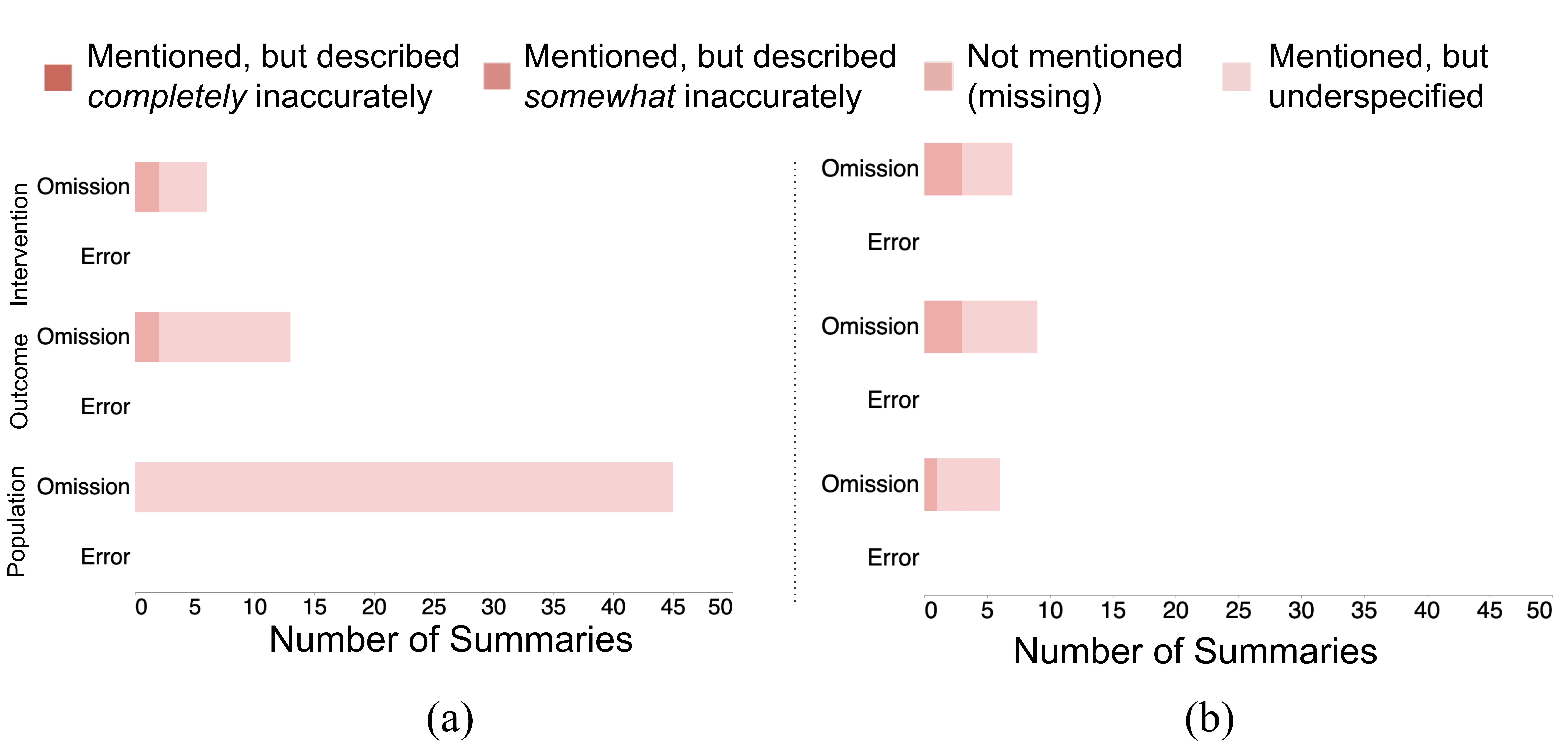}
    \vspace{-1.5em}
    \caption{Granular omissions and errors annotated in (a) technical and (b) simplified summaries. Most omissions come from underspecifying key components.}
    \label{fig:fig5_1}
\end{figure}

\begin{figure}
    \centering
    \includegraphics[width=0.45\textwidth]{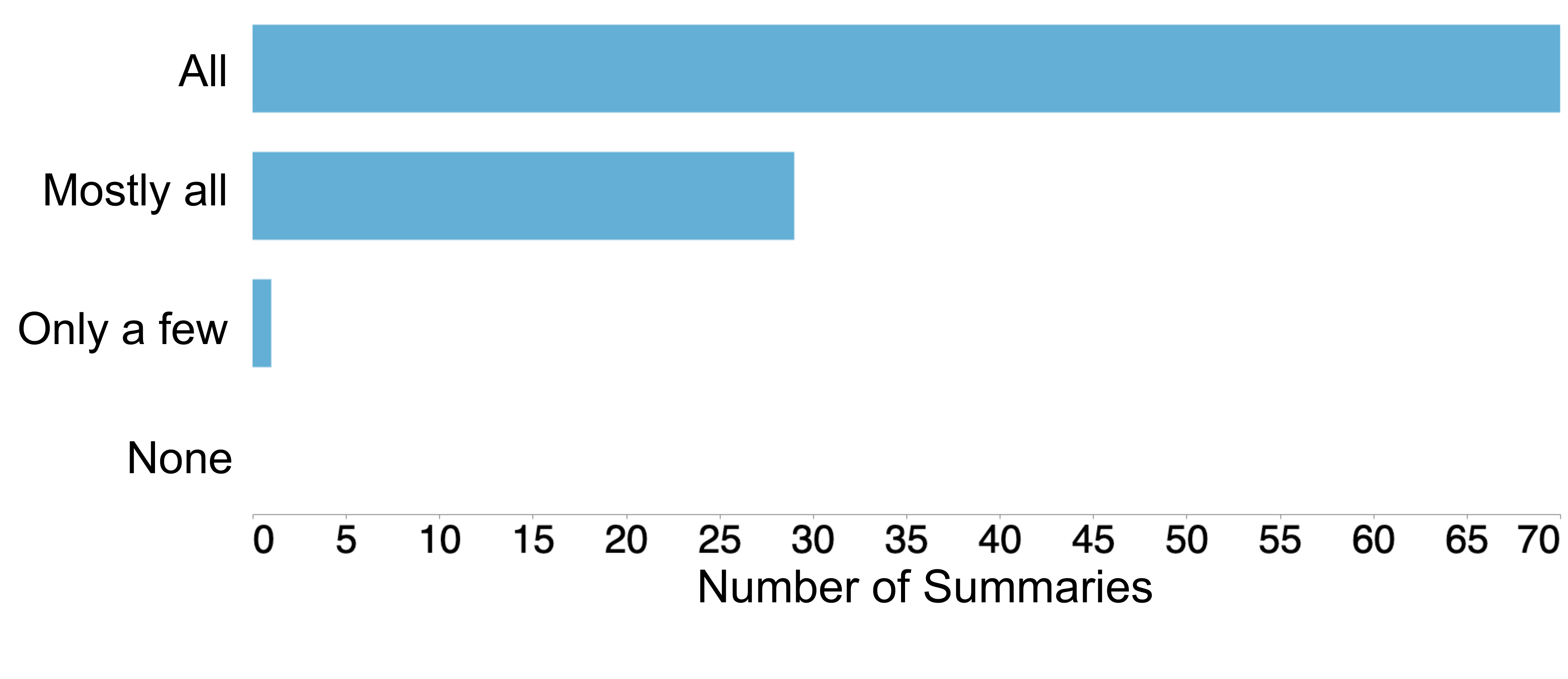}
    \vspace{-1em}
    \caption{In the simplification case, the model usually replaces complex terms with simpler ones.}
    \vspace{-0.5em}
    \label{fig:fig5_2}
\end{figure}

\vspace{-0.5em}
\paragraph{RQ4: What sort of factual mistakes does \gpt make and what are the risks?}
In RQ1, we reported that \gpt sometimes omits key information. 
Figure \ref{fig:fig5_1} characterizes the types of omissions and errors made, with respect to PICO elements. 
\gpt tends to underspecify elements in the summary more often than generating inaccuracies. Appendix \ref{sec:appendix_addl_evaluations} provides further details regarding underspecification. In the simplification task, \gpt capably simplifies most technical terms in the generated output (Figure \ref{fig:fig5_2}). 

\begin{figure}
    \centering
    \includegraphics[width=0.450\textwidth]{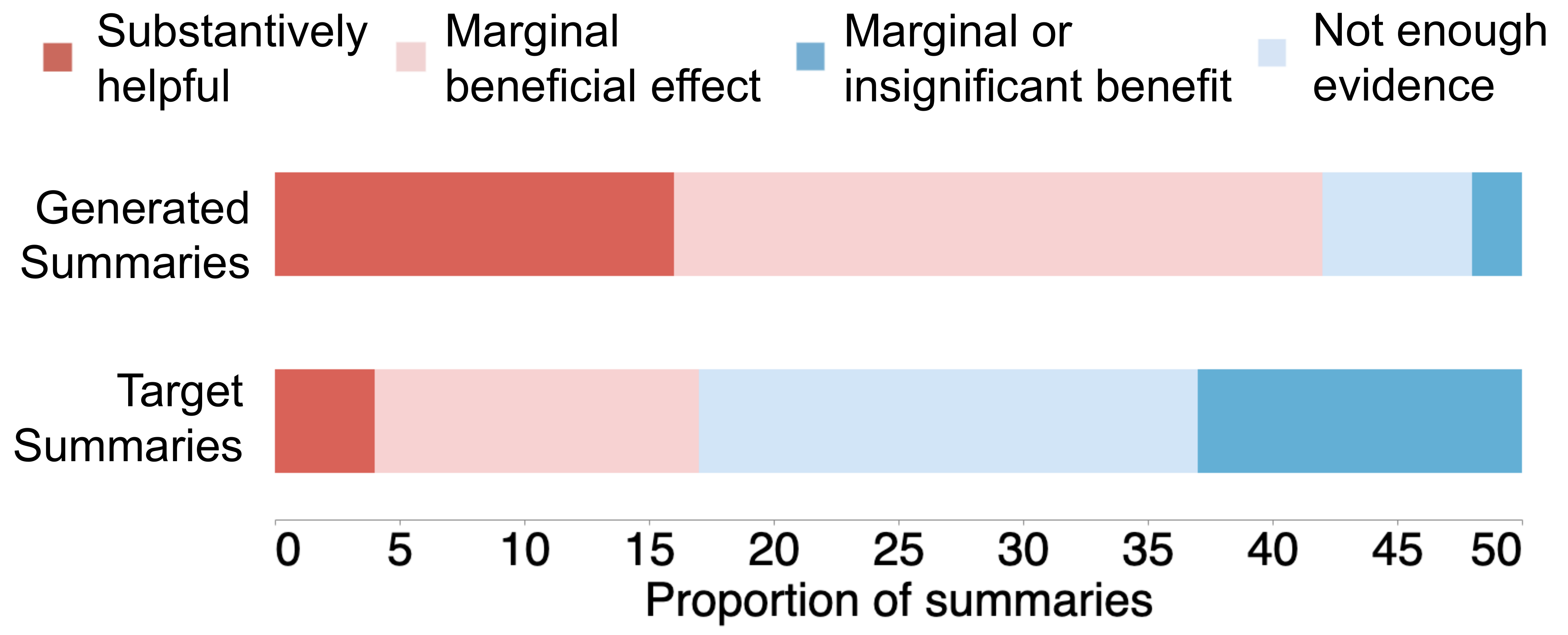}
    \caption{Proportion of summaries that are reported as beneficial in the generated summaries and the target summaries. The generated summaries tend to report beneficial effects in most of the summaries.}
    \label{fig:fig5_3}
\end{figure}

Regarding RQ3, we showed that there are often discrepancies between generated and target summaries, despite the former being supported by the inputs. 
Human-written summaries of trials may be more cautious in their conclusions. We measure the evidence strength and direction of both the target and generated summaries, and find that \gpt tends to recommend marginal or substantive beneficial effects regarding interventions in the majority of the summaries (Figure \ref{fig:fig5_3}). 

\begin{figure}[t]
    \centering
    \includegraphics[width=0.375\textwidth]{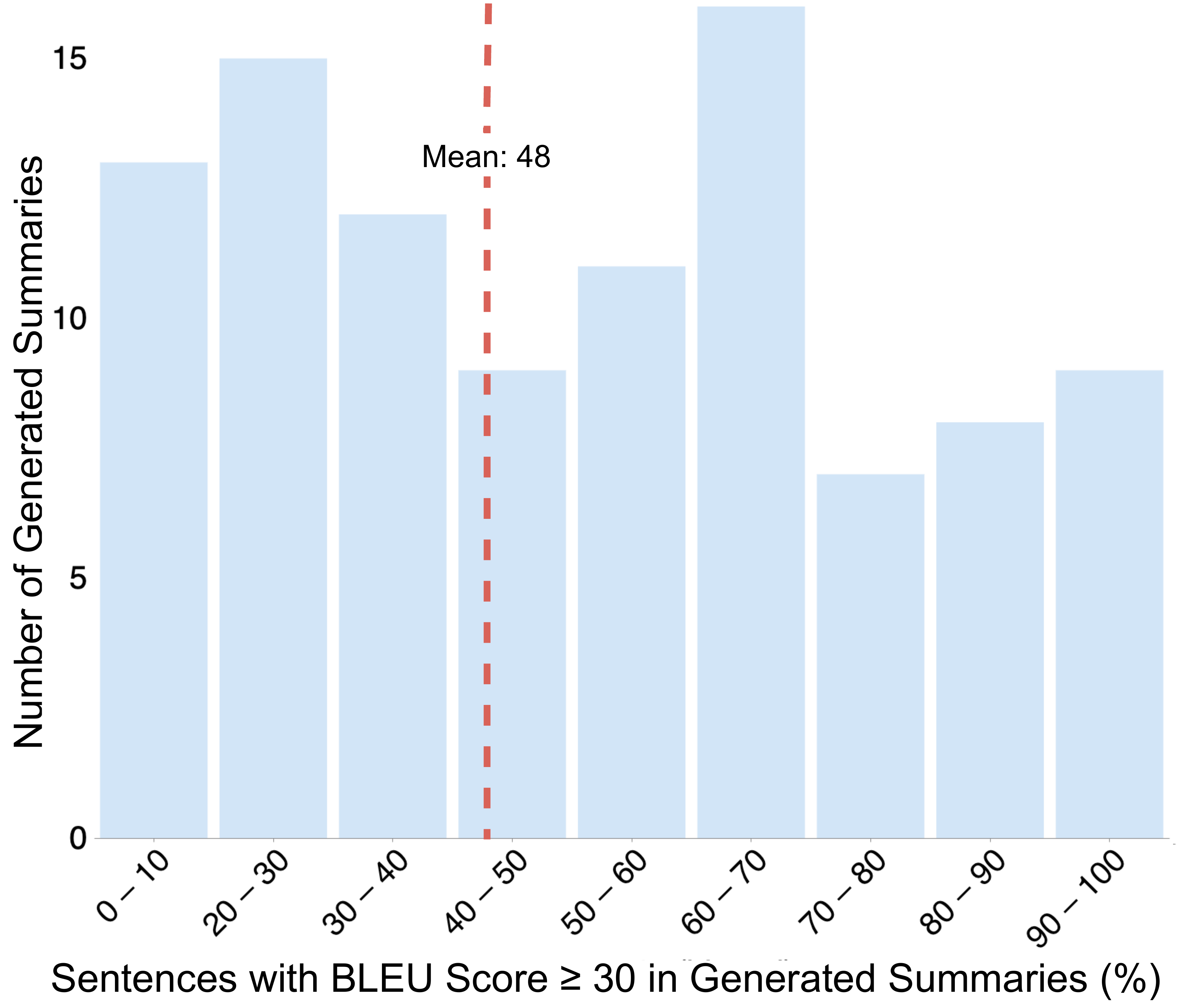}
    \caption{Percentage of sentences in the generated summaries with a BLEU score of 30 or higher, which indicates high similarity.}
    \label{fig:copied}
\end{figure}

 Overall, we find that \gpt copies frequently from inputs. 
 This results in summaries that are often faithful to the input. 
 It may also be one reason that summaries tend to have more omissions (rather than errors) in the single document case, and it may also explain how summaries in the multi-document case often disagree with the reference synopsis while also being supported by (some subset of) the inputs.
 We calculate the degree of overlap and similarity between inputs and generated summaries from \gpt for both single-document and multi-document summarization at the sentence level (Figure \ref{fig:copied}). 
\gpt often copies sentences verbatim. 
In other cases, it changes phrasings but only very slightly (see Appendix \ref{sec:appendix_addl_evaluations} for examples). 

Further, Figure \ref{fig:copied} shows how many sentences in each summary have a BLEU score of $\geq$ 30; which indicates the sentences are highly aligned. Over 70\% of the summaries have at least a quarter of the sentences copied from the input. Appendix \ref{sec:appendix_addl_evaluations} shows some examples of highly similar summaries and sentence pairs. 

\section{Related Work}

More broadly in summarization, several efforts have called for increased emphasis on human (rather than automated) evaluation of generated texts, 
increased deployment of human-centered systems for text generation evaluation \cite{Khashabi2021GENIETR}, and greater focus on building benchmarks that incorporate human preferences \cite{liang2022holistic, fabbri2021summeval}. And indeed, \citet{Goyal2022NewsSA} find that summaries produced by \gpt are often preferred by humans over alternative model outputs even when automated metrics disagree. Such findings have motivated the manual analysis we conduct for this work. As far as we know, there has not been any work that assess the degree to which GPT-3 is proficient at summarizing biomedical and clinical data in both single-document and multi-document cases.

Our analysis of summarization in the biomedical space complements recent work analyzing the question answering capabilities of such models in this domain \cite{singhal2022large, lievin2022can} and the degree to which they encode medical knowledge implicitly \cite{sung2021can}. 
Other work has 
considered using summarization of biomedical texts as assistive tools for reading \cite{August2022PaperPM}.  



\section{Conclusions} 
We evaluate the ability of \gpt to faithfully summarize and simplify medical literature.  
The expert annotations we collect indicate that \gpt performs single-document tasks quite well, but struggles with multi-document summarization.
This highlights the ability to aggregate across documents as a direction for future work. 
We release all data and annotations to facilitate such work in the medical space going forward. 

\section*{Limitations} This evaluation focussed on expert manual assessments of model outputs and their factual accuracy. Domain expertise (in medicine) was invaluable for this task, but is also expensive and therefore limited the scale of our evaluation. Consequently, all findings are derived over a modest sample (100s) of triple-annotated instances. 

Another limitation here is that we have considered only articles describing \textit{randomized control trials (RCTs)}. We focused on such articles because RCTs are the most reliable means of assessing medical interventions, and therefore inform the practice of evidence-based medicine; summarizing such articles is therefore critical to help physicians stay on top of the evidence. Moreover, RCTs provide a natural grounding with respect to factuality, given that all such trials will investigate the relative efficacy of an intervention for a particular condition (i.e., on a specific population of patients) and with respect to an outcome of interest. That said, this is restrictive by design, and our analysis has therefore excluded large swaths of other types of medical texts. 

\section*{Ethical Considerations}

In Appendix \ref{sec:annotator_info}, we note the costs of hiring domain experts for annotation.

Large language models (such as \gptwithoutspace) have been shown capable of generating concise and fluent summaries. 
But these often contain factual inaccuracies.
This poses unique risks in the domain of medicine, where inaccurate summaries of published evidence have the potential to (mis-)inform patient care. 
This work has attempted to empirically assess the tendency of models to introduce inaccuracies into summaries of medical literature by enlisting domain experts to identify and characterize omissions and errors in model generated summaries.
Understanding such issues is a first step toward designing methods to mitigate them.

While we found that \gpt appears to produce summaries of single biomedical article abstracts that are reasonably factual, relying on such outputs still poses risks, and even in this setting we would caution against trusting model outputs without further verification at present.
Moreover, we found that in the multi-document case---i.e., on the task of synthesizing evidence reported across multiple clinical trials---\gpt  struggles to provide synopses that agree with reference (expert written) summaries.
In sum, despite their ability to produce consistently plausible outputs, our view is that summaries of medical literature produced by LLMs should not yet be used to directly inform care given the risks of factual inaccuracies. 
More research is needed to better characterize the kinds of mistakes such models make, and ultimately to mitigate them. 


\section*{Acknowledgements}
This research was partially supported by National Science Foundation (NSF) grants IIS-2145479 and RI-2211954, and by the National Institutes of Health (NIH) under the National Library of Medicine (NLM) grant 2R01LM012086.

\bibliography{anthology,custom}
\bibliographystyle{acl_natbib}

\clearpage
\section*{Appendix}
\label{sec:appendix}

\renewcommand{\thesubsection}{\Alph{subsection}}

\subsection{Model details}
\label{sec:appendix_model}
We use the following parameters to prompt \gptwithoutspace: temperature = 0.7, top-p = 1.0, frequency penalty = 0.0, presence penalty = 0.0. We set our maximum token length to 1000 to avoid artificially introducing any omission errors.

\subsection{Dataset statistics}
\label{sec:dataset_statistics}
We provide some basic information about the dataset in Table \ref{table:dataset_statistics}. Because we used \gptwithoutspace, we do not have a clear idea about how the tokenization is done. To be as transparent as possible, however, we still provide the number of tokens when tokenized with SpaCy\footnote{https://spacy.io/}. Since we use \gptwithoutspace, we opt to use a tokenization scheme that focuses mainly on general English (so we did not use a specialized tokenizer for biomedical texts to replicate as similar a tokenization as possible).

\subsection{Prompts}
\label{sec:appendix_prompts}
For single-document summarization, we follow prior work to select our prompts. From \cite{Goyal2022NewsSA, August2022PaperPM}, we use the following prompts for the technical summary and the plain language summary: 

\begin{itemize}
\item Summarize the above.
\item My fifth grader asked me what this passage means: `````` [TEXT TO SIMPLIFY] '''''' I rephrased it for him, in plain language a fifth grader can understand.
\end{itemize}

 To our knowledge, there is no prior work investigating prompt constructions for multi-document summarization generally (or evidence synthesis specifically). Table \ref{table:multidoc_prompts}
  reproduces prompts we considered for this, but we ultimately used:
 
 \begin{itemize}
 \item `````` [GENERATED INPUT SUMMARIES] '''''' What does the above evidence conclude about `````` [TITLE] ''''''? 
 \end{itemize}

\begin{table}
\small 
    \centering
    \begin{tabular}{l c}
    \hline
     Prompts: \\
     \hline
        Write a meta-analysis based on the above evidence. \\
        Summarize the above evidence. \\
        Synthesize the above. \\
        
        \hline
    \end{tabular}
    \caption{Examples of prompts tried for multi-document summarization.}

    \label{table:multidoc_prompts}
\end{table}

\begin{figure}
    \centering
    \includegraphics[width=0.5\textwidth]{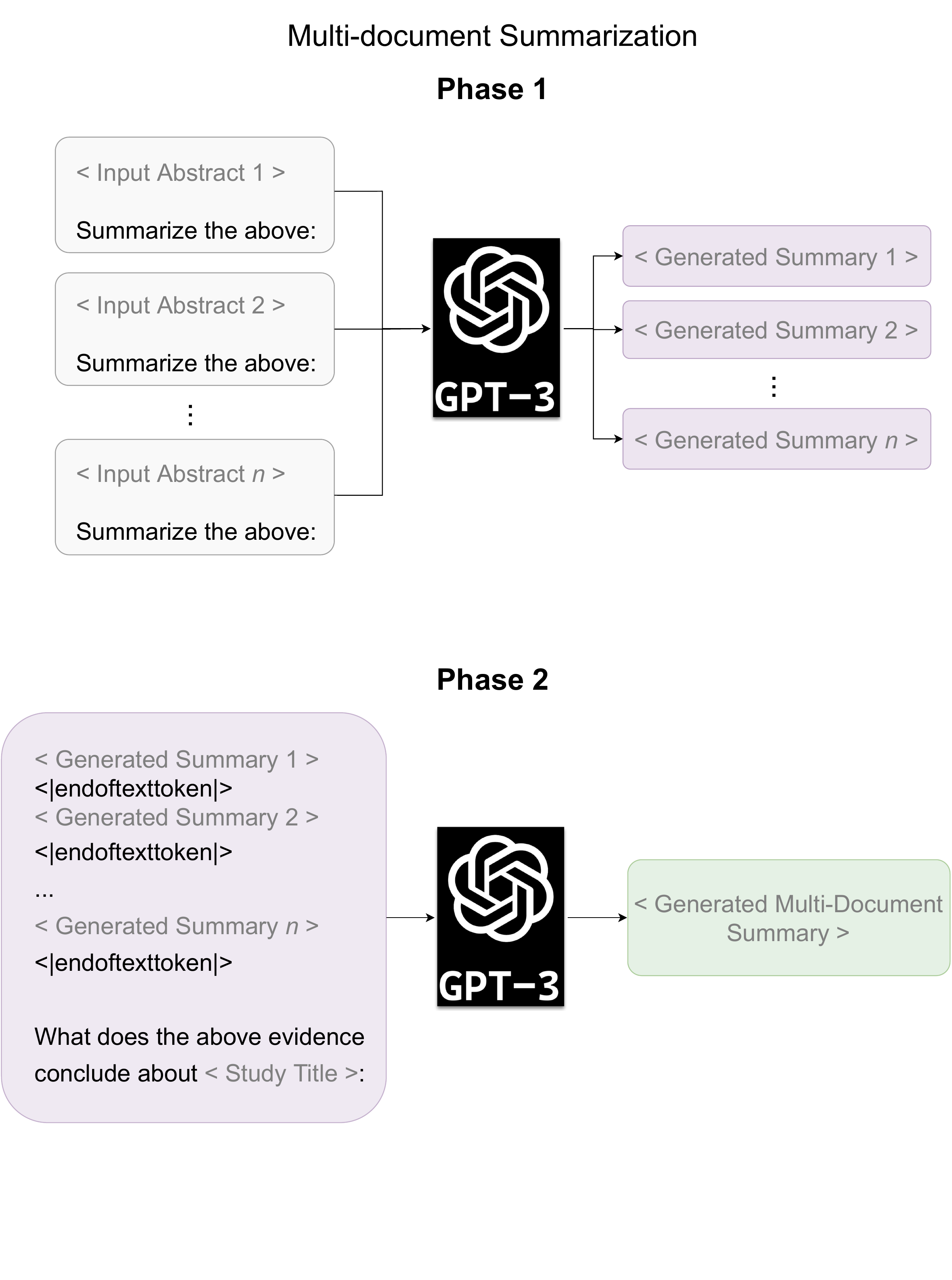}
    \caption{Input structure and prompts for the multi-document setting.}
    \label{fig:multidocphases}
\end{figure}

\begin{figure}
    \centering
    \includegraphics[width=0.5\textwidth]{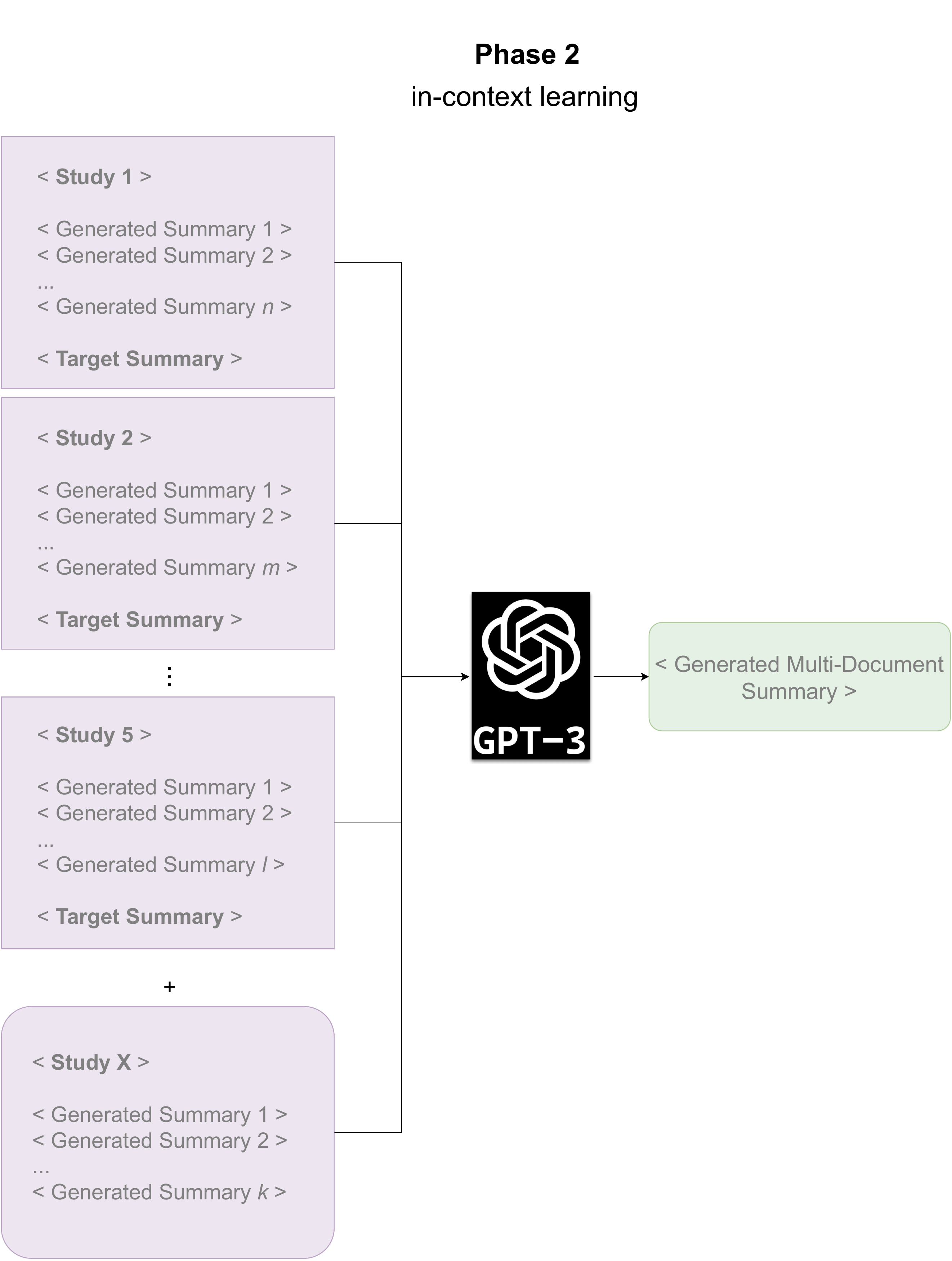}
    \caption{Adding in-context learning examples in the second step in multi-document summarization in the few-shot settings.}
    \label{fig:multidocfewshot}
\end{figure}

Figure \ref{fig:multidocphases} shows an example of the input structure and prompts we provide to \gpt in the multi-document setting. For the few-shot setting, we evaluate using up to 5 examples in context. Figure \ref{fig:multidocfewshot} shows the input structure for this setting in the second phase. 

\subsection{Annotation details}
\label{sec:annotator_info}

We calculate the inter-annotator agreement score (Cohen's kappa), which averaged 0.59 amongst all annotators. 

We also transparently reveal the cost of annotating on Upwork. The total cost of hiring 3 workers on Upwork was a little more than $\$3,700$ USD. Because annotations on a more specialized platform cost significantly more, we hired fewer annotators than one would hire on generic crowdworking websites.

Since each Upworker requested different payment amounts (which is the nature of the platform), we provide the averages per hour for the work. For the single-document case, each annotation took on average 15-20 minutes per sample, and with 100 samples, the upper-bound was 33.3 hours for the entire task per annotator. For the multi-document case, each annotation took on average 10-15 minutes per sample, and with 50 samples, the upper-bound was 12.5 hours for the entire task per annotator. Both tasks had three annotators annotating each.

\subsection{Survey details}
\label{sec:appendix_survey}

\begin{table*}
\small 
    \centering
    \begin{tabular}{lll}
    \hline
     \textbf{Type of statistic} & \textbf{Single-document} & \textbf{Multi-document} \\ \hline
        \textbf{Average number of tokens per input (all)} & 293.06 & 1451.68 \\
        \textbf{Average number of tokens per input (abstract(s) only)} & 293.06 & 1353.04 \\
        \textbf{Average number of tokens per input (study title only)} & N/A & 10.28 \\
        \textbf{Average number of tokens per input (abstract titles only)} & N/A & 88.36 \\
        \hline
    \end{tabular}
    \caption{General dataset statistics for reference. Note that in the single-document case, we only use abstracts in our zero-shot generation, so the remaining rows for anything other than abstracts only are labeled "N/A".}
    \label{table:dataset_statistics}
\end{table*}

For each data point (and for each question in the interface), the annotator first evaluates the standard summary and then evaluates the plain language summary, before completing the survey in its entirety. 
We reproduce our survey questions and the corresponding answer options. 
These include the evaluation categories that we care about: For \textbf{standard (technical) summaries}, we focus on factuality, linguistic quality, and holistic evaluation; For \textbf{plain language summaries}, we include an additional section on readability because the purpose of these is to simplify technical language such that a layperson might understand the summary.
We provide details regarding the structures of the surveys we used and our rationales behind their construction below.

\subsubsection{Single-document summarization}
In the single-document summarization case, the inputs comprise \textbf{study abstracts}, \textbf{titles}, and we also show to the user \textbf{key results}, which were automatically extracted \cite{marshall2020trialstreamer}. 
(We do not have reference summaries for these examples.)
The goal of expert evaluation was to quantify the extent to which \gpt accurately summarizes these article inputs. 
We reiterate that we consider \textbf{two} different types of summarization strategies: standard (technical) summarization and plain-language summarization. 
We reproduce the questions asked for these summary types below, which vary only slightly in their focus. 

\paragraph{Factuality} Many of our questions chosen in our taxonomy revolve around factuality since factual accuracy is extremely in domain-specific work.

\subparagraph{1. The model summary accurately conveys the key results in the input.} Given the model summary, we seek to evaluate whether the key results that are automatically extracted are reflected in the output. This is a matter of degree, so we solicit assessments rated on a Likert scale. 
\subparagraph{2. Highlight sentences in the model summary (if any) that directly contradict the input (highlight model summary on the right).} We collect additional annotations on which portions of the model summary contradict the input.
We did not further analyze these highlights here, but do release them as part of the data collected. 


\subparagraph{3. Highlight any concepts that are new in the model summary that don’t appear in the input (highlight model summary on the right).} Here the idea is to allow the annotator to mark ``hallucinated'' content in outputs (not supported by the input).

\subparagraph{4. How are details about the population described in the summary, relative to the input text?} The patient population is a critical component of clinical trials in medicine, and so it is important that summaries accurately describe this element. 
In particular we ask both whether the population is described (at all), and also the degree to which it is described \emph{accurately}. 

\subparagraph{5. How are details about the intervention described in the summary, relative to the input text?} Another key element of trials is the intervention (e.g., medicine or treatment) being evaluated. Therefore, as for study populations, we collect annotations regarding whether this is captured (and if it is captured accurately).

\subparagraph{6. How are details about the outcome (what was measured) described in the summary, relative to the input text?}
The outcome measured (e.g., mortality) is the final foundational component of trials. As in the preceding two cases, we ask annotators to assess whether this is reported upon faithfully. 

\subparagraph{7. Are there any omission(s) unrelated to the population, intervention, or outcome?} We evaluate whether the model omits any information regarding the key trial elements---population, intervention, and outcome---just described. 
For more details about types of omissions, refer to section \ref{sec:omissions}.

\subparagraph{8. Are there any errors?} We also ask whether there are any errors (in general) in the model summary. 

\paragraph{Linguistic quality} 

\subparagraph{9. The model summary is coherent, fluent, and without grammatical errors.} This is intended to capture the readability or fluency of the generated output, independent of its veracity. 

\paragraph{Holistic evaluation} Finally, we ask for a holistic evaluation of the output.

\subparagraph{10. The output is a concise, accurate, and potentially useful summary of the input.} Continuing with more holistic questions, this is intended to capture the perceived (potential) utility of generated summaries, according to the domain experts we hired as annotators. 

In the case of plain summarization, we ask the annotator to rate whether \textbf{10. The simplified text is accurate and would be understandable by a (lay) patient.} 
This effectively conveys the potential utility of automatically produced lay summaries, because the purpose of these outputs would be make medical evidence more accessible to (inexpert) patients. 


\subparagraph{11. If there was anything not elaborated or covered, feel free to leave a comment in the box.} We conclude with an open-ended text box to collect notes or thoughts not otherwise captured. 

\paragraph{Readability} For \textbf{plain language summaries}, we include a section on readability, given the focus on making evidence more digestible in this case.

\subparagraph{12. The simplified model text is less technical and more approachable, thus making it easier to understand.} This question measures the degree to which the annotator judges the model to have successfully simplified the text.

\subparagraph{13. Technical terms in the input are being substituted with simpler language in the simplified model text.} 
This is a more focussed question regarding simplification to quantify whether the model consistently swaps jargon terms for more accessible language. 

\subsubsection{Multi-document summarization} The inputs in the multi-document case comprises collections of articles describing trials, and the targets are syntheses of these (which put together the findings they report). 
We sampled these meta-reviews from previously conducted evidence syntheses, and so in this case we have target summaries, which we provide to the annotator.
We not consider simplification in the multi-document setting.

\paragraph{Factuality} We again focus on factuality of model outputs. 

\subparagraph{1. Highlight any spans in the generated summary that disagree with the target summary.} We ask for annotators to mark any explicit contradictions featured in the generated output. 

\subparagraph{2. The generated summary is supported by putting together the given summaries of the individual articles.} The core of multi-document summarization is the piecing together of multiple documents into a coherent summary that accurately reflects the inputs in aggregate. 
This question is intended to measure the degree to which the model does so.

\subparagraph{3. The generated summary agrees with the target summary.} Because we have reference (target) summaries in this case, we directly ask whether and to what degree the model generated synopsis seems to agree with this.

\subparagraph{4. Rate the degree to which the \emph{generated} summary shows the extent that there is evidence supporting the effectiveness of the intervention(s) of interest (as indicated in the studies). The \emph{generated} summary suggests...} Here we aim to assess whether the model output implies that the intervention studied in the constituent trials is supported by the findings reported within them. 

\subparagraph{5. Rate the degree to which the \emph{target} summary shows the extent that there is evidence supporting the effectiveness of the intervention(s) of interest (as indicated in the studies). The \emph{target} summary suggests...} Similarly, we ask whether the reference summary implies that the intervention in question is effective. 

\begin{table*}[h!]
    \small
    \centering
    \label{table:addl_counts}
    \begin{tabular}{ll}
        \hline
        \textbf{Type of Error} & \textbf{Number of Articles} \\ \hline
        \multicolumn{2}{l}{\textit{Population}} \\ \hline
        Omits demographic information & 0 \\
        Omits sample size & 41 \\
        Other & 1 \\ \hline
        \multicolumn{2}{l}{\textit{Intervention}} \\ \hline
        Does not describe comparator intervention & 2 \\
        Omits dosage or other important detail about administration & 1 \\
        Other & 0 \\ \hline
        \multicolumn{2}{l}{\textit{Outcome}} \\ \hline
        Omits description of specific measurements of high-level outcomes & 4 \\
        Omits one or more of multiple outcomes & 8 \\
        Other & 0 \\ \hline
    \end{tabular}
    \caption{Types of errors and the number of articles with the corresponding error, for regular summarized articles.}
    \label{table:addl_counts}
\end{table*}

\paragraph{Holistic evaluation} As above we seek to elicit an overall impression of summary accuracy and quality.

\subparagraph{6. If there was anything not elaborated or covered, feel free to leave a comment in the box.} Much like for single-document summarization, the survey provides an additional box for annotators to give information about the specific data point that was asked.

\begin{table*}[t]
    \small
    \centering
    \label{table:addl_counts_plain}
    \begin{tabular}{ll}
        \hline
        \textbf{Type of Error} & \textbf{Number of Articles} \\ \hline
        \multicolumn{2}{l}{\textit{Population}} \\ \hline
        Missing completely & 1 \\
        Missing key details (patients vs patients with depression) & 2 \\
        Inaccurate & 0 \\
        Other & 1 \\ \hline
        \multicolumn{2}{l}{\textit{Intervention}} \\ \hline
        Missing completely & 1 \\
        Missing comparator & 2 \\
        Inaccurate & 0 \\
        Other & 2 \\ \hline
        \multicolumn{2}{l}{\textit{Outcome}} \\ \hline
        Missing completely & 0 \\
        Missing part outcomes & 3 \\
        Missing key details that would be important for a lay person to know & 1 \\
        Inaccurate & 0 \\
        Other & 0 \\ \hline
    \end{tabular}
    \caption{Types of errors and the number of articles with the corresponding error, for plain summarized articles.}
    \label{table:addl_counts_plain}
\end{table*}

\subsection{Additional evaluation}
\label{sec:appendix_addl_evaluations}

\subsubsection{Few-shot}
\paragraph{Few-shot} We experimented briefly with \emph{few}-shot prompting (Appendix \ref{sec:appendix_examples}), but qualitatively this did not seem to outperform zero-shot summarization, hence our focus on evaluating the latter. 

For few-shot generation, we insert in-context training examples after the first summarization phase by concatenating the summaries and the target conclusions of inputs (see Appendix \ref{sec:appendix_prompts}). We evaluate using up to 5 shots. 

\subsubsection{Underspecified elements} 
\label{sec:omissions}

Table \ref{table:addl_counts} and Table \ref{table:addl_counts_plain} show the additional options selected when an element (e.g., population) was marked as ``underspecified'' in the survey for the technical and simplified cases, respectively.

There can be many reasons why an element could be marked underspecified. Because we try to remove as much ambiguity as possible, we opt to identify the reasons under each category (\textit{Population}, \textit{Intervention}, \textit{Outcome}) the specific reasoning. The questions we ask in both the regular and plain summarization case are both different because of the audience we address in either case. In the regular summarization case, the reader is intended to be a domain expert; in the plain summarization case, the reader is intended to be laymen, and so we alter the types of questions we ask as a result.

We find that plain summaries (Table \ref{table:addl_counts_plain}) have fewer errors than that of regular summaries (Table \ref{table:addl_counts}), whereas regular summaries have a higher number of specific omissions. However, plain summaries seem to have more omissions in areas outside of the scope of what we identify as salient omissions. We can hypothesize that given more complex language, it could be that annotators can more easily identify salient information in the text. On the other hand, there are nuances in regular summaries that cannot be extrapolated via plain summarization prompts, and instead we must use regular summaries to gather more critical information (in addition to the fact that the questions asked in the plain summarization case tends to be simpler). Although, with regular summaries, summarizing on a deeper level may result in using more convoluted language. Nonetheless, each type of prompt (regular and plain) seem to be well-suited for the task at hand; what matters is the context in which the prompt is used, and what information is needed for the user.

\subsubsection{Flan-T5}
\label{flan}
\begin{figure*}[b]
    \centering
    \includegraphics[width=\textwidth]{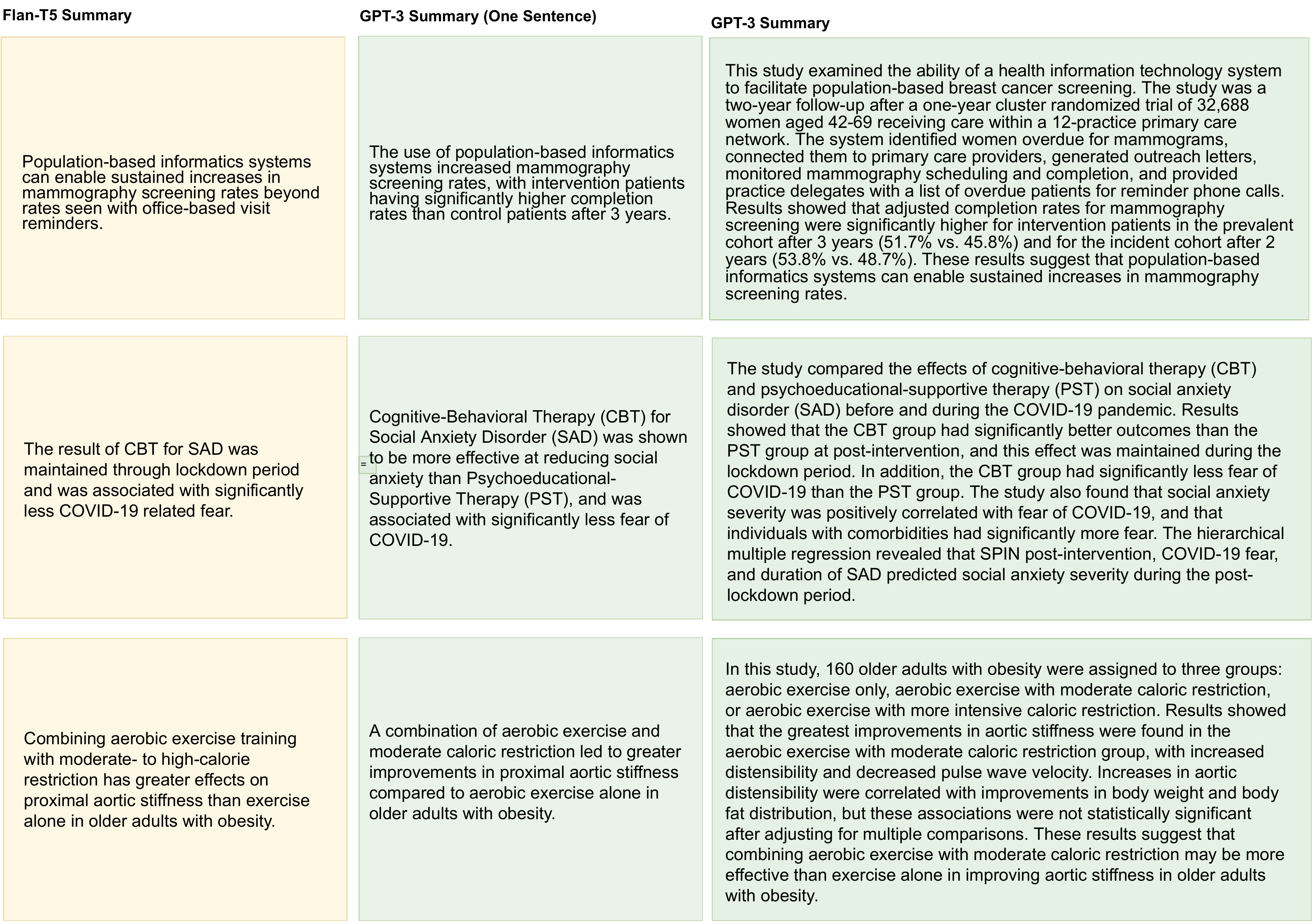}
    \caption{Sample generated summaries from Flan-T5 and GPT-3}
    \label{fig:flant5_ex}
\end{figure*}

We compared GPT-3 zero-shot results to Flan-T5 \citep{Wei2021FinetunedLM}. We find that Flan-T5 produces substantially shorter summaries (2-3 sentences on average). We provide examples of  generated summaries in Figure \ref{fig:flant5_ex}. 
Qualitatively, these seemed far worse than GPT-3 generated outputs, so we did not evaluate these further in this work.

\subsubsection{ROUGE scores}
\begin{table}[h!]
\centering
\small
\ra{1.3}
    \begin{tabular}{@{}ccc@{}} \toprule
    ROUGE-1 & ROUGE-2 & ROUGE-L \\ \midrule
    0.27 & 0.06 & 0.16 \\
    \bottomrule
    \end{tabular}
\caption{ROUGE scores on \textbf{multi-document} biomedical summaries using \gptwithoutspace}
\label{table:multidoc_rouge}
\end{table}

We provide the standard automatic metric of \texttt{ROUGE} \cite{lin-2004-rouge} to analyze multi-document summarization. We do not have \texttt{ROUGE} scores for single-document summarization since we lack ground truth data. 
However, the focus of this work is on the capability of \gpt to faithfully summarize biomedical literature (i.e., to generate accurate summaries); human experts remain the best judges of factuality. 
Noting this and prior work by \citet{Goyal2022NewsSA} make \texttt{ROUGE} scores (and other automatic metrics) rather unreliable to judge the capabilities of these large language models on summarization.

\subsubsection{Similarity}
We provide additional examples of sentences and summaries with high similarity to the input abstract. 

\begin{figure*}
    \centering
    \includegraphics[width=\textwidth]{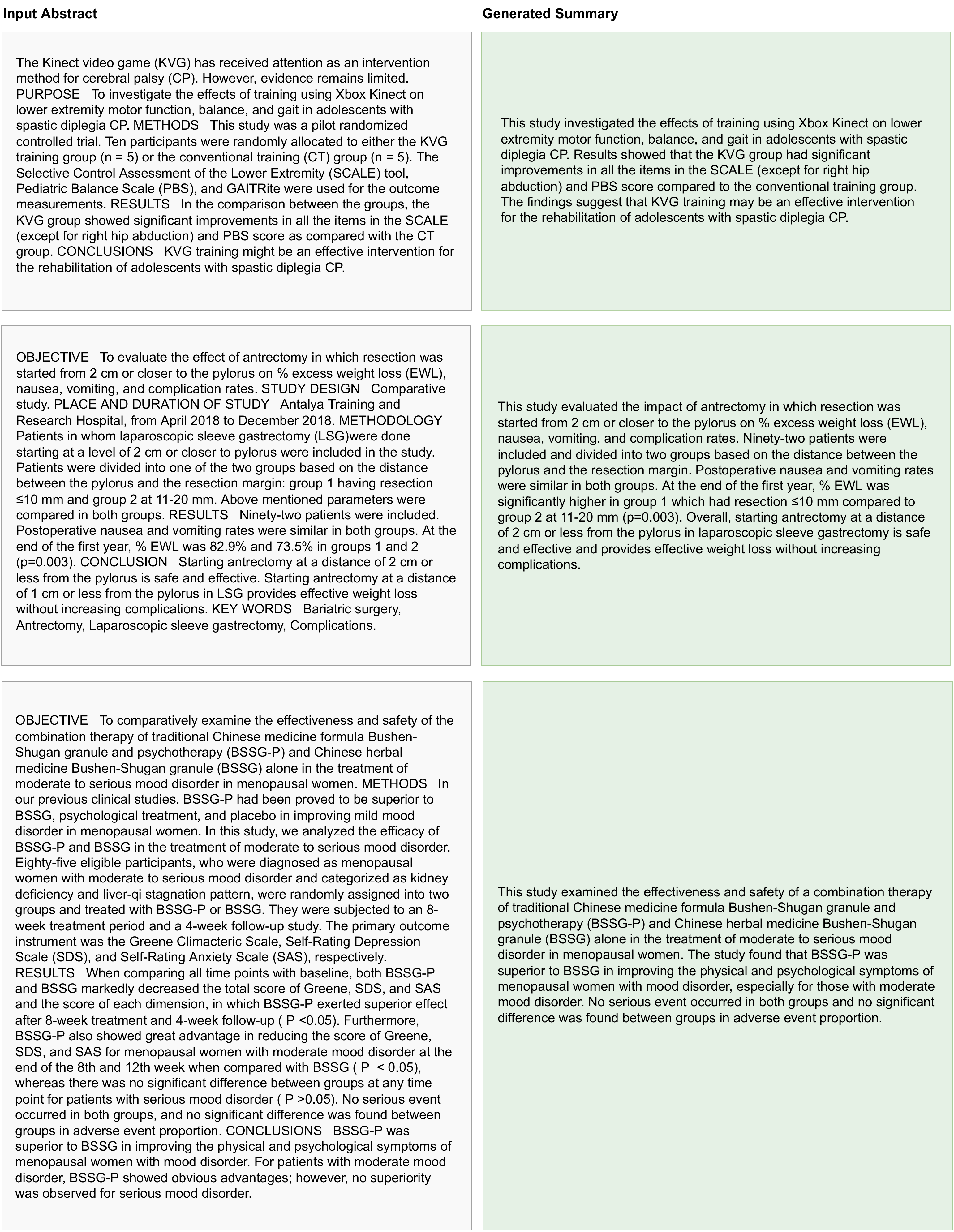}
    \caption{Examples of generated summaries where all sentences have a high similarity scores.}
    \label{fig:copied_ex}
\end{figure*}

\begin{table*}[p]
\small
    \centering
    \label{table:single_doc_sentence_pairs}
    \begin{tabular}{ p{5.6cm}  p{5.6cm} l } \toprule
        \textbf{Sentence from Abstracts} & \textbf{Sentence from Generated Summary} & \textbf{BLEU} \\ \hline
        These findings suggest that access to care and differences in treatment may be responsible for racial disparities in colorectal cancer. &  These findings suggest that access to care and differences in treatment may be responsible for racial disparities in colorectal cancer. & 100 \\ \hline
        After corrections for multiple comparisons, only PFC effects on praise and emotion strategies at post-treatment, and praise and withdrawn/depressed behavior at follow-up, maintained. & After corrections for multiple comparisons, only PFC effects on praise and emotion strategies at post-treatment, and praise and withdrawn/depressed behavior at follow-up, were maintained. & 91.93 \\ \hline
        AIM   To assess the safety and efficacy of hybrid closed-loop (HCL) insulin delivery 24/7 versus only evening and night (E/N), and on extended 24/7 use, in free-living children with type 1 diabetes. & This study aimed to assess the safety and efficacy of hybrid closed-loop (HCL) insulin delivery 24/7 versus only evening and night (E/N), and on extended 24/7 use, in free-living children with type 1 diabetes. & 91.20 \\ \hline
        We find that protocol compliance, as measured by correlations between e-cigarette use measures and cotinine levels, was only achieved in the first week of the study and declined thereafter. & The findings showed that protocol compliance, as measured by correlations between e-cigarette use measures and cotinine levels, was only achieved in the first week of the study and declined thereafter. & 90.46 \\ \hline
        CONCLUSIONS   Our findings suggest that the SERT-enriched functional network is dynamically different in ASD during processing of socially relevant stimuli. & The findings suggest that the SERT-enriched functional network is dynamically different in ASD during processing of socially relevant stimuli. & 89.96 \\ 
           \bottomrule 
    \end{tabular}
    \caption{Examples of highly extractive sentence pairs found from generated summaries for single-document summarization.}
\end{table*}\begin{table*}[p]
\small
    \centering
    \label{table:multi_sentence_pairs}
    \begin{tabular}{ p{5.6cm}  p{5.6cm} l } \toprule
        \textbf{Sentence from Abstracts} & \textbf{Sentence from Generated Summary} & \textbf{BLEU} \\ \hline
           CONCLUSIONS: Drug-induced remission of JIA-U did not persist when adalimumab was withdrawn after 1-2 years of treatment. & However, remission of JIA-U did not persist when adalimumab was withdrawn after 1-2 years of treatment. & 84.80 \\ \hline
           CONCLUSION: This study suggests that increasing the dose of inhaled steroids at the onset of an exacerbation of asthma is ineffective and should not be included in asthma self management plans. & The evidence suggests that increasing the dose of inhaled corticosteroids at the onset of an exacerbation of asthma is ineffective and should not be included in asthma self management plans. & 79.19 \\ \hline
           RESULTS: Following maternal betamethasone administration (day 2), fetal heart rate variation was reduced by 19\% and fetal body and breathing movements by 49\% and 85\%, respectively. & Dexamethasone had a greater beneficial effect, reducing fetal heart rate variation by 19\% and fetal body and breathing movements by 49\% and 85\%, respectively. & 56.71 \\ \hline
           OBJECTIVE: This study aimed to investigate the effect of endometrial injury using Pipelle catheter in the follicular phase (cycle day 5, 6, or 7) of the stimulation cycle on pregnancy rates in patients undergoing intrauterine insemination. & The evidence suggests that endometrial injury using a Pipelle catheter in the follicular phase (cycle day 5, 6, or 7) of the stimulation cycle may improve pregnancy rates in women undergoing intrauterine insemination (IUI). & 56.22 \\ \hline
           CONCLUSION: Based on these results, it is suggested that VAC has advantages when compared to the Bogota bag as a temporary closure method in the management of abdominal compartment syndrome. & Furthermore, the VAC system has advantages compared to the Bogota bag as a temporary closure method in the management of abdominal compartment syndrome. & 54.32 \\ 
           \bottomrule 
    \end{tabular}
    \caption{Examples of highly extractive sentence pairs found from generated summaries for multi-document summarization.}
\end{table*}

\subsection{Examples of generated summaries}
\label{sec:appendix_examples}
\begin{figure*}[]
    \centering
    \includegraphics[width=\textwidth]{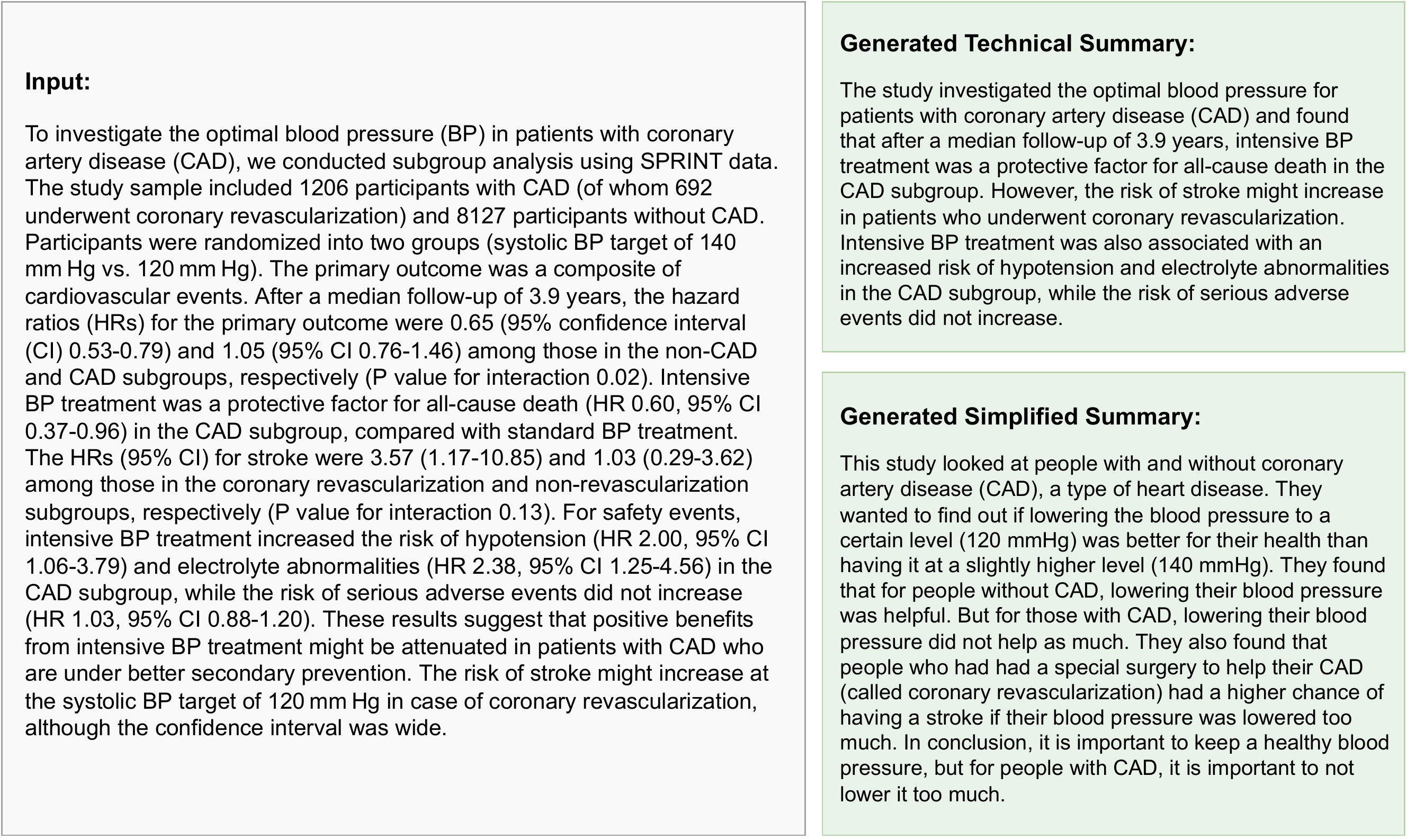}
    \caption{An example input and output (technical and simplified summaries) for the single-document summarization task.}
    \label{fig:singledocex}
\end{figure*}

\begin{figure*}[]
\centering
    \begin{varwidth}{\textwidth}
    \includegraphics[width=1.0\textwidth]{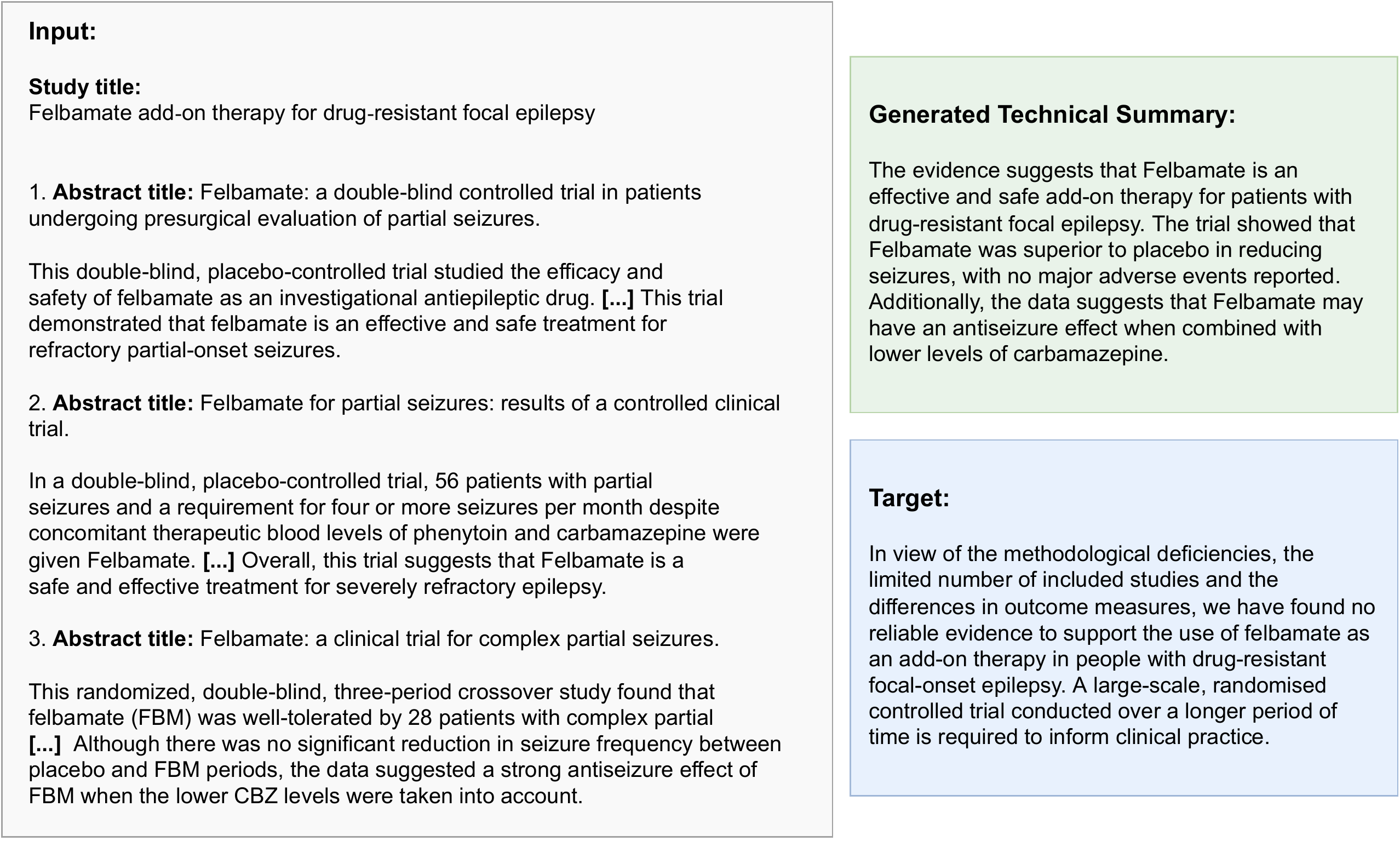}
    \caption{An example input, output, and target for the multi-document summarization task.}
    \label{fig:multidocex}
    \end{varwidth}  
\end{figure*}

We include examples of generated summaries we annotated, both standard summaries and plain language in the single and multi-document case (Table \ref{fig:multidocex}, \ref{fig:singledocex}).

\begin{figure*}[]
    \centering
    \includegraphics[width=1.0\textwidth]{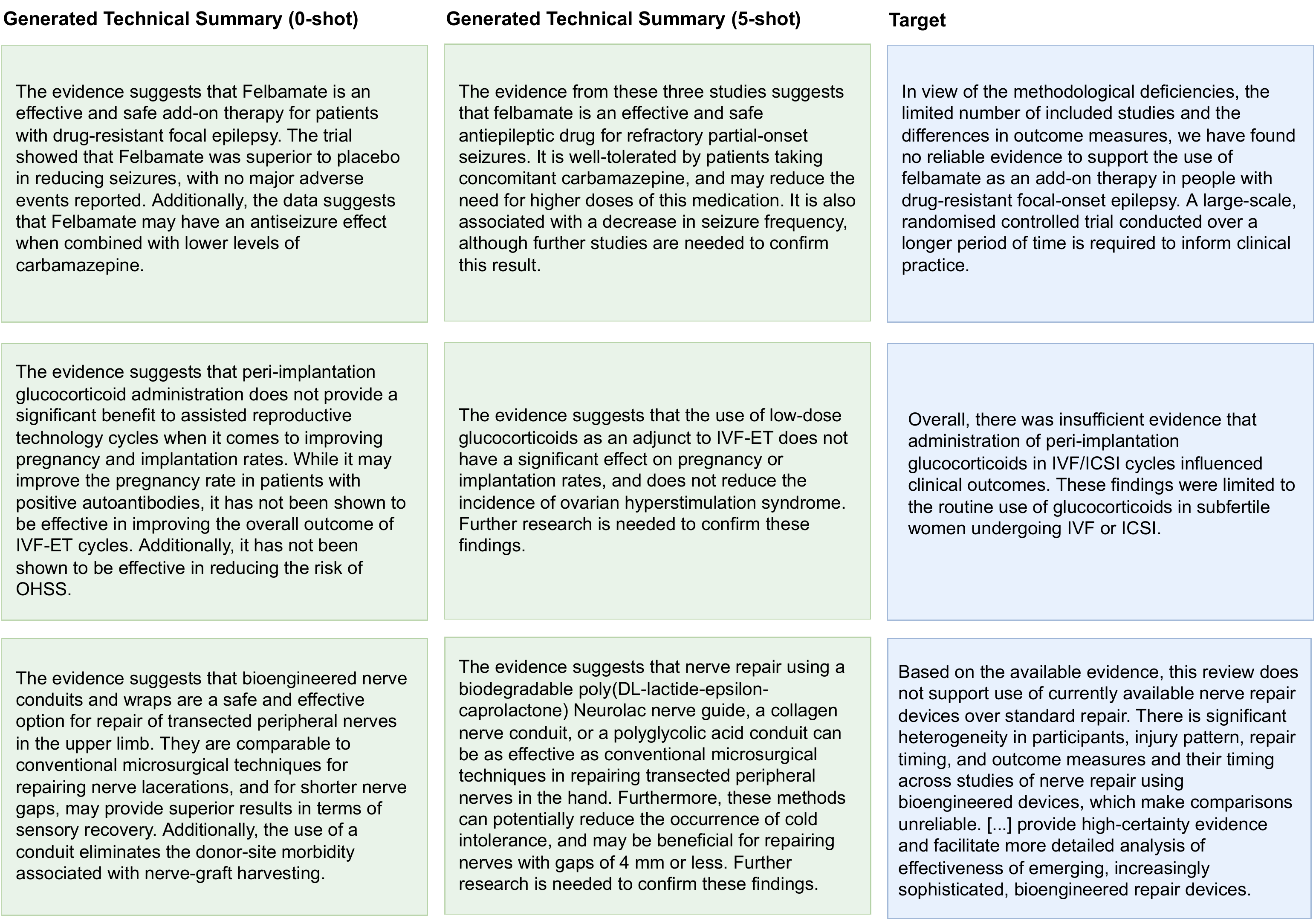}
    \caption{Examples of generated summaries in the few-shot setting and their associated target summaries}
    \label{fig:fewshot_ex}
\end{figure*}

We also provide examples of few-shot generations along with the zero-shot and target summaries for comparison (Figure \ref{fig:fewshot_ex}). Note that the few-shot examples reflect the same evidence strength and recommendation as the zero-shot examples, thus we do not evaluate them at this point.

\subsection{Additional figures}
\label{sec:survey}

\begin{table*}[p]
    \centering
    \small
    \label{crouch}
    \begin{tabular}{  l  p{5.6cm}  p{5.6cm} } \toprule
        \textbf{Evaluation Category} & \textbf{Question or Statement} & \textbf{Answer Choices} \\ \hline
            Factuality & The model summary accurately conveys the key results in the input & Strongly disagree; disagree; agree; strongly agree \\ \hline
            Factuality & Highlight sentences in the model summary (if any) that directly contradict the input (highlight model summary on the right) & Multiple tokens highlighted \\ \hline
            Factuality & Highlight any concepts that are new in the model summary that don't appear in the input (highlight model summary on the right) & Multiple tokens highlighted \\ \hline
            Factuality & How are details about the population described in the summary, relative to the input text? & The population is not mentioned (missing) in the model summary; The population is mentioned, but described completely inaccurately; The population is mentioned, but described somewhat inaccurately; The population is mentioned, and described accurately; The population is underspecified; Not applicable (N/A) \\ \hline
            Factuality & How are details about the intervention described in the summary, relative to the input text? & The intervention is not mentioned (missing) in the model summary; The intervention is mentioned, but described completely inaccurately; The intervention is mentioned, but described somewhat inaccurately; The intervention is mentioned, and described accurately; The intervention is underspecified; Not applicable (N/A) \\ \hline
            Factuality & How are details about the outcome (what was measured) described in the summary, relative to the input text? & The outcome is not mentioned (missing) in the model summary; The outcome is mentioned, but described completely inaccurately; The outcome is mentioned, but described somewhat inaccurately; The outcome is mentioned, and described accurately; The outcome is underspecified; Not applicable (N/A) \\ \hline
            Factuality & Are there any omission(s) unrelated to the population, intervention, or outcome? & No omission; Minor omission(s); Major omission(s) \\ \hline
            Factuality & Are there any errors? & No errors; Minor error; Major error \\ \hline
            Linguistic Quality & The model summary is coherent, fluent, and without grammatical errors & Strongly disagree; disagree; agree; strongly agree \\ \hline
            Holistic evaluation & The output is a concise, accurate, and potentially useful summary of the input & Strongly disagree; disagree; agree; strongly agree \\ \hline
            Holistic evaluation & If there was anything not elaborated or covered, feel free to leave a comment in the box & Free text \\
        \bottomrule
    \end{tabular}
    \caption{Questions used in our survey for annotators to evaluate standard summaries}
\end{table*}

\begin{table*}[p]
\small
    \centering
    \label{crouch}
    \begin{tabular}{  l  p{5.6cm}  p{5.6cm} } \toprule
        \textbf{Evaluation Category} & \textbf{Question or Statement} & \textbf{Answer Choices} \\ \hline
            Factuality & The simplified model text accurately conveys the key results in the input & Strongly disagree; disagree; agree; strongly agree \\ \hline
            Factuality & Highlight sentences in the input (if any) that directly contradict the simplified model text (highlight input on the right) & Multiple tokens highlighted \\ \hline
            Factuality & Highlight any concepts that are new in the simplified model text that don't appear in the input (highlight model summary on the right) & Multiple tokens highlighted \\ \hline
            Factuality & How are details about the population described in the simplified model text, relative to the input text? & The population is not mentioned (missing) in the simplified model text; The population is mentioned, but described completely inaccurately; The population is mentioned, but described somewhat inaccurately; The population is mentioned, and described accurately; The population is underspecified; Not applicable (N/A) \\ \hline
            Factuality & How are details about the intervention described in the simplified model text, relative to the input text? & The intervention is not mentioned (missing) in the simplified model text; The intervention is mentioned, but described completely inaccurately; The intervention is mentioned, but described somewhat inaccurately; The intervention is mentioned, and described accurately; The intervention is underspecified; Not applicable (N/A) \\ \hline
            Factuality & How are details about the outcome (what was measured) described in the simplified model text, relative to the input text? & The outcome is not mentioned (missing) in the simplified model text; The outcome is mentioned, but described completely inaccurately; The outcome is mentioned, but described somewhat inaccurately; The outcome is mentioned, and described accurately; The outcome is underspecified; Not applicable (N/A) \\ \hline
            Factuality & Are there any omission(s) unrelated to the population, intervention, or outcome? & No omission; Minor omission(s); Major omission(s) \\ \hline
            Factuality & Are there any errors? & No errors; Minor error; Major error \\ \hline
            Linguistic Quality & The simplified text is coherent, fluent, and without grammatical errors & Strongly disagree; disagree; agree; strongly agree \\ \hline
            Holistic evaluation & The simplified text is accurate and would be understandable by a (lay) patient & Strongly disagree; disagree; agree; strongly agree \\ \hline
            Holistic evaluation & If there was anything not elaborated or covered, feel free to leave a comment in the box & Free text \\
        \bottomrule
    \end{tabular}
    \caption{Questions used in our survey for annotators to evaluate simplified model summaries}
\end{table*}

\begin{table*}[p]
    \centering
    \small
    \label{crouch}
    \begin{tabular}{  l  p{5.6cm}  p{5.6cm} } \toprule
        \textbf{Evaluation Category} & \textbf{Question or Statement} & \textbf{Answer Choices} \\ \hline
            Readability & The simplified model text is less technical and more approachable, thus making it easier to understand. & Strongly disagree; disagree; agree; strongly agree \\ \hline
            Readability & Technical terms in the input are being substituted with simpler language in the simplified model text. & None at all; Only a few; Mostly all; All \\
        \bottomrule
    \end{tabular}
    \caption{Additional questions used in our survey for annotators to evaluate simplified model summaries}
\end{table*}

\begin{table*}[p]
\small
    \centering
    \label{crouch}
    \begin{tabular}{  l  p{5.6cm}  p{5.6cm} } \toprule
        \textbf{Evaluation Category} & \textbf{Question or Statement} & \textbf{Answer Choices} \\ \hline
           Factuality & Highlight any spans in the generated summary that disagree with the target summary & Multiple tokens highlighted \\ \hline
           Factuality & The generated summary is supported by putting together the given summaries of the individual articles & Strongly disagree; disagree; agree; strongly agree \\ \hline
           Factuality & The generated summary agrees with the target summary & Strongly disagree; disagree; agree; strongly agree \\ \hline
           Factuality & Rate the degree to which the *generated* summary shows the extent that there is evidence supporting the effectiveness of the intervention(s) of interest (as indicated in the studies). The *generated* summary suggests… & There is not enough evidence to draw any meaningful conclusions; The intervention has a marginal or insignificant comparative benefits; The intervention may have a marginal beneficial effect; The intervention is substantively helpful \\ \hline
           Factuality & Rate the degree to which the *target* summary shows the extent that there is evidence supporting the effectiveness of the intervention(s) of interest (as indicated in the studies). The *target* summary suggests… & There is not enough evidence to draw any meaningful conclusions; The intervention has a marginal or insignificant comparative benefits; The intervention may have a marginal beneficial effect; The intervention is substantively helpful \\ \hline
           Holistic Evaluation & If there was anything not elaborated or covered, feel free to leave a comment in the box & Free text \\ \hline
           \bottomrule 
    \end{tabular}
    \caption{Questions used in our survey for annotators to evaluate multi-document model summaries}
\end{table*}

\end{document}